\documentclass[conference,final]{IEEEtran}

\ifCLASSOPTIONcompsoc
  \usepackage[nocompress]{cite}
\else
  \usepackage[nospace]{cite}
\fi

\usepackage{comment}

\ifCLASSINFOpdf
  \usepackage[pdftex]{graphicx}
\else
  \usepackage[dvips]{graphicx}
\fi
\usepackage{nameref}

\usepackage[cmex10]{amsmath}
\usepackage{listings} 

\lstdefinelanguage{pseudocode}{
    morekeywords={class, do, loop, function, call, repeat},
    sensitive=false, 
    morecomment=[l]{//}, 
    morestring=[b]" 
}

\usepackage[caption=false,font=footnotesize]{subfig}
\usepackage{url}
\usepackage{color}
\usepackage{enumerate}
\usepackage{tikz}
\usepackage[utf8]{inputenc}
\usepackage{pdflscape}
\usepackage{pgfplots} 
\usepackage{pgfgantt}
\pgfplotsset{compat=newest} 
\pgfplotsset{plot coordinates/math parser=false}
\newlength\fwidth

\begin{document}
%
\title{NeoPhysIx: An Ultra Fast 3D Physical Simulator \\
as Development Tool for AI Algorithms}


\author{
\IEEEauthorblockN{
J\"orn Fischer\IEEEauthorrefmark{1} and
Thomas Ihme\IEEEauthorrefmark{2}}
\IEEEauthorblockA{Robotics Institute\\
Mannheim University of Applied Sciences, 68163 Mannheim, Germany\\
\IEEEauthorrefmark{1}j.fischer@hs-mannheim.de,
\IEEEauthorrefmark{2}t.ihme@hs-mannheim.de}
}
\maketitle

\begin{abstract}

Traditional AI algorithms, such as Genetic Programming and Reinforcement Learning, often require extensive computational resources to simulate real-world physical scenarios effectively. While advancements in multi-core processing have been made, the inherent limitations of parallelizing rigid body dynamics lead to significant communication overheads, hindering substantial performance gains for simple simulations.
 
This paper introduces NeoPhysIx, a novel 3D physical simulator designed to overcome these challenges. By adopting innovative simulation paradigms and focusing on essential algorithmic elements, NeoPhysIx achieves unprecedented speedups exceeding 1000x compared to real-time. This acceleration is realized through strategic simplifications, including point cloud collision detection, joint angle determination, and friction force estimation.
 
The efficacy of NeoPhysIx is demonstrated through its application in training a legged robot with 18 degrees of freedom and six sensors, controlled by an evolved genetic program. Remarkably, simulating half a year of robot lifetime within a mere 9 hours on a single core of a standard mid-range CPU highlights the significant efficiency gains offered by NeoPhysIx. This breakthrough paves the way for accelerated AI development and training in physically-grounded domains.

\end{abstract}

\bstctlcite{IEEEexample:BSTcontrol}

\section{INTRODUCTION}
To develop effective robot controllers, optimize robots' planning behavior, or enable offline learning techniques such as Reinforcement Learning, very fast and stable physical simulators are essential. These simulators must handle 3D rigid bodies, as well as various types of sensors and actuators, to accurately model complex robotic systems \cite{Fischer2010, Fischer2004, Pilat2004, Watkins1992, fischer2001_1, fischer2001_2, fischer2002_1}.

Over the years, simulation algorithms have been significantly optimized to deliver faster results \cite{Jorge}. Many simulators now meet the real-time performance criterion. However, with the 4.0 GHz clock frequency barrier for CPUs in place \cite{Naffziger2011}, further improvements in simulation speed have stagnated.

Real-time simulation is particularly important for the gaming industry, where the most advanced simulators are optimized to meet real-time requirements even in complex scenarios. Despite these improvements, extremely fast simulation speeds are still required for applications such as evolving learning neuro-controllers, executing real-time planning in complex environments, or optimizing robot construction parameters.

Simulators like PhysX (NVIDIA), ODE (Open Dynamics Engine), Havok, and Bullet are fine-tuned to deliver data at rates exceeding real-time. However, achieving a frame rate that is a multiple of real-time often requires substantial software restructuring or implementing unconventional solutions, such as using separate threads for each computed scene. These optimizations reflect the primary focus on meeting game physics requirements rather than the unique needs of robotics.

Projects involving Artificial Intelligence, where algorithms require the capability to test behaviors across diverse scenarios, demand specialized simulators. These simulators must not only meet real-time performance requirements but also deliver results at the highest possible speeds.

The following sections introduce a new simulator, NeoPhysIx, capable of achieving up to one million frames per second on a single thread of a standard Intel i5 processor. Key methods that enable this speedup, such as point cloud collision detection, joint angle calculation, and friction force estimation, will be discussed in detail.

\section{Related Work}

Numerous commercial and open-source rigid body simulators exist, each developed with significant effort to ensure accurate and consistent physical behavior. This raises the question: why introduce yet another simulator? What specific applications might this new simulator target?

To address these questions, we can categorize applications that rely on simulation as follows:

\begin{enumerate}
\item High-precision physical simulations for predictive purposes (e.g., architecture, medical simulations, crash testing).
\item Real-time simulations with lower precision (e.g., game engines, virtual reality).
\item Simulations faster than real time (e.g., weather forecasting, tsunami warning systems).
\item Ultra-fast simulations for AI applications such as planning, evolutionary algorithms, and reinforcement learning.
\end{enumerate}

Traditionally, planning algorithms, reinforcement learning, and evolutionary algorithms have relied on simulation environments developed for the gaming industry, using simulators like ODE, Bullet, Havok, and PhysX \cite{Fischer2010, Fischer2004}. When simulating moderately complex robots, these tools typically achieve speedups ranging from 1 to 20 times real time. This performance is often insufficient, resulting in longer scenario testing times, smaller population sizes, and prolonged waiting times for results. In particular, evolving complex neural controllers with self-modifying synapses is challenging and has been demonstrated in only a few specific cases \cite{Soltoggio}.

One study on the Webots simulator \cite{Michel2004} mentions a "fast simulation" mode that is reported to run up to 300 times faster than real time. However, the study lacks details on the robot’s architecture and the required hardware specifications to achieve this acceleration, making the reported data insufficient as a benchmark. Another study \cite{Zahedi2008} failed to reproduce these results: “This could not be validated with the examples provided in the evaluation version (Ver. 5, Mac OS 10.5.4, 2.5 GHz Dual Core, 4 GB RAM). The maximum speed achieved was approximately seven times faster than real time."

One notable project from Cornell University focuses on optimizing the PhysX simulation framework specifically for neuro-evolution. In this setup, a four-legged robot with eight joints was simulated at speeds 264 times faster than real time. This significant speedup was achieved using specialized NVIDIA hardware, disabling visualization, and “forcing multiple threads managing scene models onto local threads or processor cores” \cite{Gicklhorn}. By comparison, NeoPhysIx can accelerate the same model to 2,235 times real time.

The following sections describe the functioning of NeoPhysIx. Sections \ref{GP} and \ref{results} demonstrate the simulation framework’s capabilities in a genetic programming context, where a genetic programming algorithm discovers solutions for controlling a complex walking machine within approximately six months of accumulated robot runtime. The final sections present conclusions and an overview of the NeoPhysIx API commands.

\section{Computation Principles}
\label{cp}

Since rigid bodies can be viewed as approximations of extremely stiff objects, there is no single physically accurate model for their behavior. As a result, various solution methods are available, and the goal is to establish consistent rules that provide stable and realistic behavior at the highest possible speed.

Simulation speed is influenced by the algorithm’s complexity, the simulated object’s complexity, and the program’s structure. Multi-threading may be beneficial when communication overhead is minimal and when other processor cores are available. In applications like planning, reinforcement learning, or evolutionary algorithms, a multi-threaded framework could enable testing robots in separate environments, each in a dedicated thread. However, NeoPhysIx runs on a single thread to simplify its structure and enhance performance.

\begin{figure}[tb]
\begin{lstlisting}
function main():
   do
      repeat n times:
         call simulateStep()

      // The calculation of the scene 
      // is too time-consuming to be 
      // carried out for each simulation
      // step.

      call render()
   loop

class PhysicsEngine:
   function simulateStep():
      // Here, physics is computed
      ...
      call move()

class RigidBody:
  ...

class Robot extends RigidBody:
  ...

// User-defined functions
class SixLeggedRobot extends Robot:
   function SixLeggedRobot():
      // Robot construction code 

   function move():
      // Reads sensor values
      // and moves joints
      ...
\end{lstlisting}
\caption{Overview of the software structure. In fast mode, the function simulateStep() is called one million times while the renderer is called only once. To implement a custom robot, users create a class derived from the Robot class. In the constructor, the robot’s geometry is defined, and the move function is called during simulation.}
\label{figurelabel}
\end{figure}

NeoPhysIx was developed using Microsoft Visual C++ with 64-bit compilation and optimization flags enabled. Profiling tools were employed throughout development to eliminate unnecessary code and optimize performance. The graphical output is minimal and platform-independent, relying on OpenGL. The same codebase also compiles on 64-bit Debian Linux, achieving comparable speed.

To achieve realistic behavior with minimal computational power, the methods described in the following sections are applied.

\subsection{Point Cloud Collision}
\label{pointcloudcolision}

The robot model comprises a set of mass points, which collectively represent the robot's structure. These mass points serve two main purposes: calculating the center of gravity (CoG) and determining the robot's momentum. Although a triangular mesh is defined along these mass points for visual representation, it currently serves no role in collision detection. However, future implementations may incorporate this mesh to enable more precise collision detection between different robots.

Since detecting collisions using triangle intersections is computationally intensive, we adopt a simplified approach that considers only the interaction between the robot's mass points and a predefined height map representing the environment. This approach is implemented through a straightforward single-loop algorithm. In this algorithm, the $x$-$y$ coordinates of each mass point are scaled by a factor of 10, converting them into integer indices. These indices are used to look up the associated height value in a 2D array, representing the height map as a look-up table. 

The choice of a scaling factor of 10, when using meters, kilograms, and seconds (MKS units), yields a height map resolution of $0.1$ meters, sufficient for modeling basic indoor and outdoor environments. For scenarios involving smaller robots or environments that require higher precision, this scaling factor may be increased accordingly.

\subsection{Joint Angle Determination}

A key distinction between this system and conventional physics engines is the treatment of the robot as a single rigid body. Relative movements between different robot segments are computed directly, bypassing the need for force-based calculations. This approach is valid only when the motors are sufficiently powerful to achieve the desired angles within the given time frame, and when movements occur at a pace slow enough relative to the simulation's timestep.

NeoPhysIx updates the physical state of the simulation every 10\,ms of simulated time. Calculating new joint angles at each frame introduces significant computational overhead. To optimize performance, joint angle calculations are typically spaced over multiple timesteps. Empirical tests indicate that updating joint angles every 80\,ms yields realistic behavior while maintaining computational efficiency. This interval is adjustable, allowing for tuning based on specific simulation requirements to balance accuracy with performance.

\subsection{Friction Force Estimation}

As discussed in Section~\ref{pointcloudcolision}, even utilizing a look-up table for collision detection demands significant computational resources. Conventionally, friction forces are computed at each contact point and then summed for integration. However, NeoPhysIx employs an alternative approach by modeling friction as the maximum static friction and formulating the friction computation as an optimization problem. Specifically, at each time step, the algorithm minimizes the distance between a contact point at time $t$ and the corresponding contact point at time $t+1$, allowing for both displacement and rotation of the robot. 

For example, in a 2D case involving both displacement and rotation, the following equation describes the new contact point position:

\begin{equation}
\begin{pmatrix}
x^{t+1} \\
z^{t+1}
\end{pmatrix}
=
\begin{pmatrix}
\cos(\alpha) & -\sin(\alpha) \\
\sin(\alpha) & \cos(\alpha)
\end{pmatrix}
\begin{pmatrix}
x^{t} \\
z^{t}
\end{pmatrix}
+
\begin{pmatrix}
a_1 \\
a_2
\end{pmatrix}
\end{equation}

Here, $x^{t}$ and $z^{t}$ are the coordinates of the contact point at time $t$, $\binom{a_1}{a_2}$ represents the displacement vector, and $\alpha$ is the rotation angle necessary to adjust the robot’s position such that the distance to the same contact point at time $t+1$ is minimized. This nonlinear equation is then approximated via a first-order Taylor expansion:

\begin{equation}
\begin{pmatrix}
x^{t+1} \\
z^{t+1}
\end{pmatrix}
\approx
\begin{pmatrix}
1 & -a_3 \\
a_3 & 1
\end{pmatrix}
\begin{pmatrix}
x^{t} \\
z^{t}
\end{pmatrix}
+
\begin{pmatrix}
a_1 \\
a_2
\end{pmatrix}
\end{equation}

The resulting optimization problem is then solved using a least squares approach:

\begin{IEEEeqnarray}{rCl}
\frac{\partial}{\partial a_j}
\sum_i \Big[&&\left(x_i^t - a_3 z_i^t + a_1 - x_i^{t+1}\right)^2\IEEEnonumber\\
+ &&\left(x_i^t a_3 + z_i^t + a_2 - z_i^{t+1}\right)^2\Big] = 0
\end{IEEEeqnarray}

\noindent where $i$ indexes all contact points existing in both time steps. Solving this results in a system of three linear equations, solvable when more than two contact points are involved:

\begin{IEEEeqnarray}{rrlCl}
\sum_i&2&(x_i^t - a_3 z_i^t + a_1 - x_i^{t+1}) &=& 0 \\
\sum_i&2&(x_i^t a_3 + z_i^t + a_2 - z_i^{t+1}) &=& 0 \\
\sum_i&\Big[2&(x_i^t - a_3 z_i^t + a_1 - x_i^{t+1})(-z_i^t)&&\IEEEnonumber\\
&+2&(x_i^t a_3 + z_i^t + a_2 - z_i^{t+1})(x_i^t)\Big] &=& 0
\end{IEEEeqnarray}

Here, $\binom{a_1}{a_2}$ is the displacement vector, and $a_3 \approx \sin(\alpha)$ is an approximation for the sine of the rotation angle. Despite these simplifications, the resulting friction model yields remarkably realistic outcomes. For greater accuracy, forces inducing rotations around both the $x$- and $z$-axes should also be considered. Further refinement may be achieved by weighting each term in the sum by the force pressing the contact point against the ground.

\section{The Application Programming Interface}

NeoPhysIx is meticulously engineered to provide a comprehensive and user-friendly application programming interface (API) that streamlines the process of robot simulation. The API prioritizes intuitive design principles, enabling researchers and developers to efficiently construct and manipulate virtual robotic systems. 

At the core of NeoPhysIx's API lies the `robot` class, which serves as the foundational building block for constructing robot models. This class encapsulates a suite of essential functions required for robot simulation, simplifying the development workflow.  

\subsection{Body Definition and Customization}

NeoPhysIx offers a diverse library of primitive body types to facilitate the creation of intricate robotic structures. Users can readily generate various geometric shapes, including mass points, boxes, cylinders, spheres, and rays, through dedicated function calls for each body type. This modular approach allows for rapid prototyping and experimentation with different robot configurations.

Furthermore, each body possesses an optional color attribute, enabling users to visually distinguish individual components within the simulated environment. This feature proves invaluable for debugging and understanding complex robotic systems.

\subsection{Assembly and Joint Definition}

The `connectBodies()` function serves as the cornerstone for assembling the constituent bodies into a cohesive robot structure. This function accepts a list of body objects as input, establishing connections between them based on user-defined parameters. 

To endow robots with mobility and articulation, NeoPhysIx provides the `createJoint()` function. This function enables the creation of hinge joints between connected bodies, allowing for controlled movement and manipulation within the simulated environment. The joint parameters, such as axis of rotation and range of motion, can be meticulously defined to accurately replicate real-world robotic behavior.

This comprehensive API empowers users to construct sophisticated robot models with ease, fostering a streamlined and efficient development process for robotics research and simulation.

\begin{figure}[tb]
  \centering
  \subfloat[Sphere]{
    \begin{tikzpicture}[y=0.80pt, x=0.80pt, yscale=-0.500000, xscale=0.500000, inner sep=0pt, outer sep=0pt]
  \path[draw=black,line join=miter,line cap=butt,line width=0.800pt]
    (221.8618,105.2303) -- (150.0434,126.5098) -- (189.9425,110.0182) -- cycle;
  \path[draw=black,line join=miter,line cap=butt,line width=0.800pt]
    (221.8618,105.2303) -- (150.5754,152.5772) -- (149.5114,125.9778);
  \path[draw=black,line join=miter,line cap=butt,line width=0.800pt]
    (150.0434,152.0453) -- (192.2035,170.6649) -- (222.3938,105.2303) --
    (249.5252,171.1968) -- (192.4695,170.6648);
  \path[draw=black,line join=miter,line cap=butt,line width=0.800pt]
    (249.9242,170.9308) -- (289.4243,153.6412) -- (222.2608,105.3633) --
    (290.6213,127.8398) -- (247.9292,108.8212) -- (222.5268,105.4963);
  \path[draw=black,line join=miter,line cap=butt,line width=0.800pt]
    (149.7774,126.5098) -- (124.5080,164.8130) -- (125.5720,197.7962) --
    (182.2287,222.5337) -- (260.1650,221.7357) -- (314.4277,197.2642) --
    (315.4917,163.7490) -- (290.0893,127.8398);
  \path[draw=black,line join=miter,line cap=butt,line width=0.800pt]
    (289.4243,153.6412) -- (314.6937,197.3972);
  \path[draw=black,line join=miter,line cap=butt,line width=0.800pt]
    (260.6969,221.7357) -- (249.5252,170.6649);
  \path[draw=black,line join=miter,line cap=butt,line width=0.800pt]
    (192.1350,170.6971) -- (182.4947,222.2677);
  \path[draw=black,line join=miter,line cap=butt,line width=0.800pt]
    (150.5089,152.3778) -- (125.5720,197.7962) -- (150.0434,234.5034) --
    (190.4745,251.5270) -- (249.5252,252.0590) -- (290.4883,233.9714) --
    (314.8267,197.2643);
  \path[draw=black,line join=miter,line cap=butt,line width=0.800pt]
    (260.6969,221.2037) -- (250.4562,251.7930);
  \path[draw=black,line join=miter,line cap=butt,line width=0.800pt]
    (181.9627,221.7357) -- (189.5435,251.3940) -- (219.2019,257.3789) --
    (250.3232,252.0590);
  \path[draw=black,line join=miter,line cap=butt,line width=0.800pt]
    (222.9258,186.0925) -- (314.9597,197.1312);
  \path[draw=black,line join=miter,line cap=butt,line width=0.800pt]
    (289.4243,153.6412) .. controls (289.6016,144.7748) and (289.7790,132.3334) ..
    (289.9563,127.0418);
  \path[draw=black,line join=miter,line cap=butt,line width=0.800pt]
    (220.4275,182.5893) -- (225.2237,188.4670);
  \path[draw=black,line join=miter,line cap=butt,line width=0.800pt]
    (225.2237,182.8244) -- (220.1924,188.2319);
  \path[cm={{0.98997,0.14125,-0.14125,0.98997,(0.0,0.0)}},fill=black,line
    join=miter,line cap=butt,line width=0.800pt] (279.1235,147.4470) node[above
    right, rotate=-8] (text3484) {\footnotesize radius};

\end{tikzpicture}
    \label{body_sphere}}%
    \quad
  \subfloat[Box]{
    \begin{tikzpicture}[y=0.80pt, x=0.80pt, yscale=-0.500000, xscale=0.500000, inner sep=0pt, outer sep=0pt]
  \path[draw=black,line join=miter,line cap=butt,line width=0.800pt]
    (239.7894,199.5461) -- (279.6637,161.1765) -- (400.4150,161.1765) --
    (362.4216,198.7938) -- cycle;
  \path[draw=black,line join=miter,line cap=butt,line width=0.800pt]
    (239.4133,199.9223) -- (239.7894,318.4166) -- (362.0454,319.1689) --
    (361.6693,199.1699);
  \path[draw=black,line join=miter,line cap=butt,line width=0.800pt]
    (362.4216,319.5451) -- (400.7912,283.0564) -- (400.7912,161.9289);
  \path[draw=black,line join=miter,line cap=butt,line width=0.759pt]
    (316.6722,238.6515) -- (321.0648,244.4294);
  \path[draw=black,line join=miter,line cap=butt,line width=0.815pt]
    (321.3940,238.7312) -- (316.3311,244.3050);
  \path[fill=black] (279.8016,336.0937) node[above right] (text4006) {\footnotesize width};
  \path[fill=black] (288.7499,264.8610) node[above right] (text4010) {\footnotesize $x$, $y$, $z$};
  \path[cm={{0.70138,-0.71279,0.71279,0.70138,(0.0,0.0)}},fill=black]
    (27.5081,505.2239) node[above right, rotate=41] (text4014) {\footnotesize depth};
  \path[cm={{0.03119,-0.99951,0.99951,0.03119,(0.0,0.0)}},fill=black]
    (-272.0580,240.4360) node[above right, rotate=90] (text4018) {\footnotesize height};

\end{tikzpicture}
    \label{body_box}}\\%
  \subfloat[Cylinder]{
    \begin{tikzpicture}[y=0.80pt, x=0.80pt, yscale=-0.500000, xscale=0.500000, inner sep=0pt, outer sep=0pt]
  \path[draw=black,line join=miter,line cap=butt,line width=0.800pt]
    (121.4165,209.6098) -- (150.1466,181.2558) -- (240.0518,88.3412) --
    (278.0452,88.3412) -- (188.1400,180.8796) -- (216.3530,209.8449) --
    (306.2582,117.6827) -- (278.7976,88.3412);
  \path[draw=black,line join=miter,line cap=butt,line width=0.800pt]
    (306.3522,117.7767) -- (307.0105,158.3093) -- (216.7291,248.9668) --
    (215.9768,209.8449);
  \path[draw=black,line join=miter,line cap=butt,line width=0.800pt]
    (188.8924,180.8796) -- (150.5228,180.8796);
  \path[draw=black,line join=miter,line cap=butt,line width=0.800pt]
    (121.5575,209.3277) -- (121.5575,250.8477) -- (151.2751,279.0606) --
    (188.8924,279.2393) -- (216.7748,248.9440);
  \path[fill=black] (169.5379,222.1646) node[above] (text4101) {\footnotesize $x_1,y_1,z_1$};
  \path[draw=black,line join=miter,line cap=butt,line width=0.761pt]
    (167.7415,226.6981) -- (172.2636,232.3393);
  \path[draw=black,line join=miter,line cap=butt,line width=0.800pt]
    (172.4969,226.8073) -- (167.4656,232.2148);
  \path[cm={{0.93343,0.35875,-0.35875,0.93343,(0.0,0.0)}},fill=black]
    (254.8950,171.3972) node[above, rotate=-21.5] (text4191) {\footnotesize radius};
  \path[cm={{0.71514,-0.69898,0.69898,0.71514,(0.0,0.0)}},fill=black]
    (50.6747,351.8873) node[above, rotate=45] (text4101-5-1) {\footnotesize length};
  \path[draw=black,line join=miter,line cap=butt,even odd rule,line width=0.800pt]
    (169.9894,229.5023) -- (216.7857,248.2550);
  \path[fill=white,rounded corners=0.0000cm] (235.0000,110.9336) rectangle
    (264.2857,128.4336);
  \path[fill=black] (256.6026,127.1191) node[above] (text4101-4) {\footnotesize $x_2,y_2,z_2$};
  \begin{scope}[shift={(89.27223,-92.42896)},shift={(0,0)}]
    \path[draw=black,line join=miter,line cap=butt,line width=0.761pt]
      (167.7415,226.6981) -- (172.2636,232.3393);
  \end{scope}
  \begin{scope}[shift={(89.27223,-92.42896)},shift={(0,0)}]
    \path[draw=black,line join=miter,line cap=butt,line width=0.800pt]
      (172.4969,226.8073) -- (167.4656,232.2148);
  \end{scope}

\end{tikzpicture}
    \label{body_cylinder}}%
    \qquad
  \subfloat[Ray]{
    \begin{tikzpicture}[y=0.80pt, x=0.80pt, yscale=-0.500000, xscale=0.500000, inner sep=0pt, outer sep=0pt]
  \path[draw=black,line join=miter,line cap=butt,line width=0.800pt]
    (233.8455,96.5833) -- (100.3165,225.8564);
  \path[fill=black] (113.5001,232.6842) node[above right] (text4101-6) {\footnotesize $x_1,y_1,z_1$};
  \path[fill=black] (220.1380,95.4717) node[below left] (text4101-5-3)
    {\footnotesize $x_2,y_2,z_2$};
  \path[draw=black,line join=miter,line cap=butt,line width=0.800pt]
    (231.5911,93.0721) -- (236.3873,98.9498);
  \path[draw=black,line join=miter,line cap=butt,line width=0.800pt]
    (236.3873,93.3072) -- (231.3560,98.7146);
  \path[draw=black,line join=miter,line cap=butt,line width=0.800pt]
    (97.8337,222.9175) -- (102.6299,228.7952);
  \path[draw=black,line join=miter,line cap=butt,line width=0.800pt]
    (102.6299,223.1526) -- (97.5986,228.5601);

\end{tikzpicture}
    \label{body_ray}}%
  \caption{The NeoPhysIx simulator provides a simple API for creating and connecting various body types.}
  \label{bodies}
\end{figure}
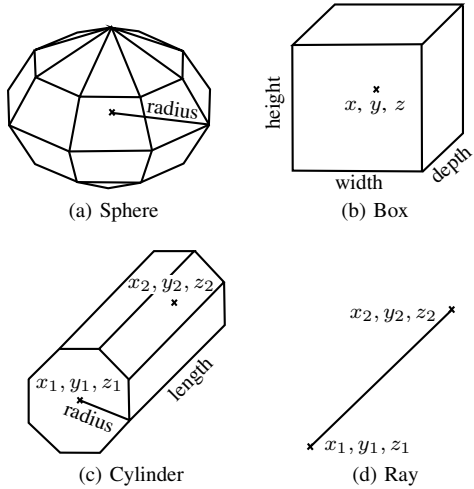

\subsection{Simplifying Robot Construction for Faster Simulation}
\label{Simplify}

To further optimize simulation speed, users can employ the \lstinline{simplifyMode()} function, which modifies the body shapes to simpler forms before creation. For instance, this feature can represent a cylindrical shape as a box, or in some cases, reduce it to one or two points. Such simplifications can significantly enhance performance. The \lstinline{simplifyMode(mode)} function accepts different modes for varying levels of simplification, as outlined below:

\addvspace{2mm}
\noindent $
mode
\begin{cases}
= 0 \text{: No simplification}\\
= 1 \text{: Convert cylinders and spheres to boxes}\\
= 2 \text{: Convert cylinders and spheres to two points}\\
\ge 3 \text{: Convert cylinders and spheres to a single point}
\end{cases}
$

\section{The Evolution of the GP Robot Controller}
\label{GP}

To evaluate the functionality and stability of NeoPhysIx, we applied an optimization scenario using Genetic Programming (GP). GP was tasked with evolving a compact machine language program to control a six-legged walking robot. The fitness function guiding this evolution was straightforward: "Walk as far as possible."

\begin{figure}[tb]
\centering
\includegraphics[width=4.5cm]{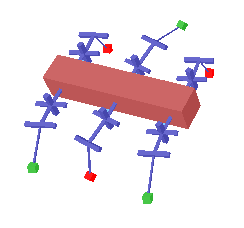}
\caption{Screenshot of the simulated robot, equipped with foot contact sensors and 18 degrees of freedom.} 
\label{hexapod}
\end{figure}

\begin{table}[tb]
\renewcommand{\arraystretch}{1.3}
\caption{Machine Code Interpreter Commands}
\label{table_machinecodeinterpreter}
\centering
\begin{tabular}{c c c}
\hline
\bfseries Opcode & \bfseries Command & \bfseries Function of xxxxx \\
\hline
000xxxxx & move motor left  & motor identifier \\           
001xxxxx & move motor right & motor identifier \\ 
010xxxxx & read sensor      & sensor identifier \\
011xxxxx & load             & direct value \\ 
100xxxxx & jump if true     & jump address \\
101xxxxx & not              & not in use \\
110xxxxx & inc              & not in use \\
111xxxxx & dec              & not in use \\
\hline
\end{tabular}
\end{table}

Table~\ref{table_machinecodeinterpreter} presents the commands implemented in the machine code interpreter. Each opcode is an 8-bit value; the upper 3 bits designate the command, while the remaining 5 bits specify a command-related value, such as the motor or sensor number. This interpreter contains a single accumulator. For example, the "load" command stores a direct value in the accumulator, and the "read sensor" command assigns a sensor value to it. The "jump if true" command enables a conditional jump to one of 32 memory addresses if the accumulator is non-zero. Additionally, the commands "not," "inc," and "dec" enable the programmer to manipulate the accumulator by inverting its bits, incrementing, or decrementing its value. These commands are essential for creating loops and conditional jumps.

To maintain simplicity, the evolutionary algorithm uses mutation as its sole variation operator, as crossover would only be practical with a hierarchical command structure. Starting from randomly generated programs, each individual is evaluated by simulating two minutes of robot activity in NeoPhysIx. A small machine code interpreter interfaces with the simulator’s actuators and sensors to simulate the controller. Each generation, the best 50 of 100 individuals are selected, duplicated, and mutated to form the next generation, followed by a new evaluation period.

\section{RESULTS}
\label{results}

This section presents the findings from our extensive robot lifetime evaluation, spanning approximately six months of simulated operation. The simulation exhibited remarkable stability throughout the experiment, consistently yielding effective controllers for the robotic agent.  

\begin{figure}[tb]
\centering
\setlength\fwidth{0.8\linewidth}
\input{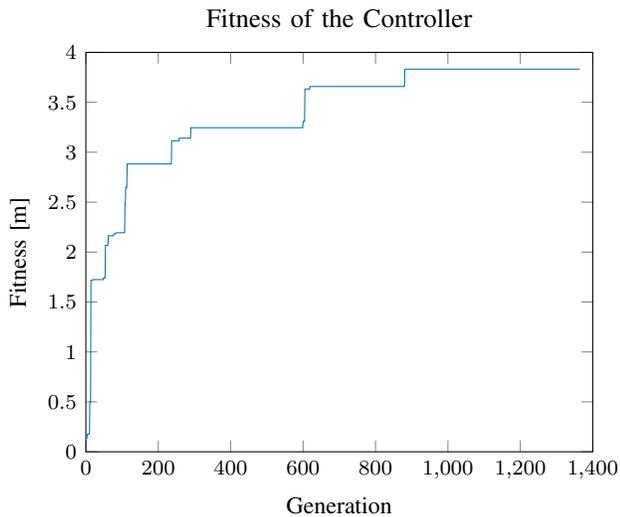}
\caption{It is shown the best fitness, which is the longest walking distance, of all robots in their generation. Each generation contains 100 robots with their controller programs. Each controller is tested 2 minutes of robot lifetime. 200 minutes x 1,364 generations results in 189 days which is more than half a year.}
\label{learningCurve}
\end{figure}

Figure \ref{learningCurve} illustrates the evolution of performance over generations. Notably, the simulation was conducted on a single thread of an Intel i5 processor operating at 2.4 GHz, demonstrating the efficiency of our approach. The computational time required to simulate two minutes of robot lifetime across 100 robots, spanning 1,364 generations, amounted to approximately nine hours. This translates to an impressive simulation duration of 189 days, effectively representing over half a year of simulated robot lifespan.

The optimization process successfully generated a machine language program that resulted in an unconventional sideways walking gait for the robot.  This behavior demonstrates the algorithm's ability to discover novel solutions and adapt to the complexities of the environment. Importantly, no memory leaks were observed during the nine-hour runtime, indicating the robustness and stability of our implementation. The simulation remained operational throughout this period and was ultimately terminated manually due to time constraints. 

The successful completion of this extensive simulation provides compelling evidence for the efficacy of our approach in developing robust and adaptable robotic controllers.  Further analysis of the generated machine language programs and their corresponding behaviors will provide valuable insights into the learning process and pave the way for future improvements.




\section{CONCLUSION}

This paper presents a novel physical simulator, termed NeoPhysIx, which incorporates simplified yet robust simulation principles. The architecture of NeoPhysIx and the methodologies employed for key functionalities such as point cloud collision detection, joint angle determination, and friction force estimation were meticulously described in previous sections. 

The efficacy of NeoPhysIx was rigorously validated through a comprehensive Genetic Programming (GP) scenario. A complex robotic system with 18 degrees of freedom and six integrated sensors was simulated within the NeoPhysIx environment. Over an extended period exceeding half a year of simulated robot lifetime, the simulator facilitated the evolution and testing of more than 100,000 distinct controller programs. This extensive experimentation demonstrated the capability of NeoPhysIx to handle intricate robotic systems and complex evolutionary algorithms.

The successful implementation of NeoPhysIx establishes a new paradigm in physical simulation, offering a unique platform for advancing research in areas such as robot learning, evolution, planning, and controller testing. 

\section{FUTURE WORK}

While NeoPhysIx demonstrates significant potential, several avenues for improvement and expansion exist.  A key area of focus is the refinement of the current friction model to enhance its accuracy and realism.  Furthermore, incorporating functionalities for simulating robot-robot collisions will significantly broaden the scope of applications. 

Expanding sensor capabilities by integrating camera sensors will provide richer environmental perception for simulated robots, enabling more sophisticated control strategies. To facilitate wider adoption and accessibility, a Python wrapper for NeoPhysIx is planned. This wrapper will enable seamless integration with existing Python ecosystems, allowing for efficient simulation execution across diverse operating systems such as Linux, Windows, and MacOS.

Currently, evolutionary and neural algorithms implemented within NeoPhysIx primarily serve as proof-of-concept demonstrations. To fully realize the simulator's potential, integrating state-of-the-art methods in these domains is crucial. This will empower researchers to explore cutting-edge control strategies and push the boundaries of robotic intelligence within the robust and versatile NeoPhysIx environment. 

\addtolength{\textheight}{-5.5cm}

\section*{APPENDIX}
To get an idea of which commands the API offers to be able to construct a robot we give a short overview here:

\subsubsection{createPoint}
A mass point is defined by its coordinates ($x$, $y$, $z$) and its mass ($m$). In addition a color (red, green, blue) using floating point values between 0 and 1 may be defined. Building a robot out of simple mass points is supposed to lead to the greatest speedup possible. The body ID of the mass point is given back.
\lstinline{bodyID createPoint(x,y,z,m,r,g,b);}

\subsubsection{createCylinder}
A cylinder (fig.~\ref{body_cylinder}) is created with two coordinates ($x_1$, $y_1$, $z_1$ and $x_2$, $y_2$, $z_2$), a radius ($rr$) and a mass ($m$). The color (red, green, blue) is optional. The function provides the body ID of the cylinder.
\lstinline{bodyID createCylinder(x1,y1,z1, x2,y2,z2,rr,m,r,g,b);}

\subsubsection{createBox}
A box (fig.~\ref{body_box}) consists of 8 points and 12 triangles. The center of the box is at $x$, $y$, $z$ and it has a length ($l$), a width ($w$), a height ($h$) and a mass ($m$). The color (red, green, blue) is optional. The function gives back the body ID of the box. 
\lstinline{bodyID createBox(w,h,d,x,y,z,m,r,g,b);}

\subsubsection{createSphere}
The center of the sphere (fig.~\ref{body_sphere}) is at $x$, $y$, $z$ coordinate. It has a radius ($rr$) and a mass ($m$). The color (red, green, blue) is optional. The function gives the body ID of the sphere back.
\lstinline{bodyID createSphere(x,y,z,rr,m,r,g,b);}

\subsubsection{createRay}
A ray (fig.~\ref{body_ray}) does not collide. It is defined by a start point $x_1$, $y_1$, $z_1$ and an end point $x_2$, $y_2$, $z_2$. The function provides the body ID.
\lstinline{bodyID  createRay(x1,y1,z1,x2,y2,z2,r,g,b);}

\subsubsection{simplifyMode}
This function may be used before the definition of a body. It enables the
user to switch between different levels of detail. e.\,g. by creating a box instead of a cylinder. This may speedup the simulation enormously. The method for simplification may be chosen according to section~\ref{Simplify}. 
\lstinline{void  simplifyMode(int mode);}

\subsubsection{connectBodies}
To generate more complex rigid bodies out of mass points, boxes, cylinders and spheres, the \lstinline{connectBodies()} function enables the programmer to glue these basic bodies together. 
\lstinline{void  connectBodies(id1,id2);}

\subsubsection{createJoint}
To create a joint between two bodies the \lstinline{createJoint()} function has to be called
with the body ids, the anchor point at $an_x$, $an_y$, $an_z$ relatively to the coordinate frame of body 1 and the direction $a_x$, $a_y$, $a_z$ of the joint axis as parameters. This function provides the joint ID.
\lstinline{jointID  createJoint(id1,id2,anX,anY,anZ,aX,aY,aZ);}

\subsubsection{getMaxBodyID}
This function returns the number of BodyIDs available.
\lstinline{bodyID getMaxBodyID();}

\subsubsection{moveBody}
This function translates the into a position $t_x$, $t_y$, $t_z$ in relation to the coordinate system used for construction.
\lstinline{void moveBody(id,tx,ty,tz);}

\subsubsection{rotateBody}
This function rotates a body about body axis $x$ with angle $\alpha$, then about axis $y$ with angle $\beta$ and finally about the $z$-axis with angle $\gamma$ following the RPY concept. The rotation direction follows the right-hand rule for each axis.
\lstinline{void rotateBody(id, alpha, beta, gamma);}

\subsubsection{finalizeConstruction}
This function has to be called when the construction is finished.
\lstinline{int finalizeConstruction();}


\IEEEtriggeratref{10}

\bibliographystyle{IEEEtran}

\bibliography{NeoPhysIxSimulator}

\end{document}